\documentclass[conference]{IEEEtran}
\IEEEoverridecommandlockouts
\usepackage{cite}
\usepackage{amsmath,amssymb,amsfonts,bm}
\usepackage{algorithmic}
\usepackage{graphicx}
\usepackage{mathrsfs}
\usepackage{placeins}
\usepackage{textcomp}
\usepackage{float}
\usepackage{xcolor}
\usepackage{array}
\usepackage{booktabs}
\usepackage{multirow}
\def\BibTeX{{\rm B\kern-.05em{\sc i\kern-.025em b}\kern-.08em
    T\kern-.1667em\lower.7ex\hbox{E}\kern-.125emX}}
\begin{document}

\title{PEPS: Quantum-Inspired Reinforcement Learning for Coherent Reasoning Traces in LLMs\\
}

\author{
    Venkat Margapuri, Garik Kazanjian, Naren Kosaraju \\
    Department of Computing Sciences, Villanova University \\
    \{vmargapu, gkazanji, nkosaraj\}@villanova.edu
}

\maketitle

\begin{abstract}
Large Language Models (LLMs) often struggle with maintaining coherent multi-step reasoning traces, particularly in tasks that require a structured logical flow. This work introduces a quantum-inspired approach to address the challenge by incorporating a fidelity-based reward derived from Projected Entangled Pair States (PEPS) into Proximal Policy Optimization. Unlike prior approaches that use direct supervision or contrastive objectives, the proposed method guides learning through structural consistency, offering a novel approach to enforce global coherence in generated reasoning traces. The proposed framework is evaluated using multiple coherence-determining metrics on diverse datasets such as GSM8K, StrategyQA, and EntailmentBank spanning arithmetic, intuitive, and entailment-based reasoning. Results show that the proposed quantum-inspired approach offers significant improvements over supervised, contrastive, and pretrained baseline approaches, highlighting the effectiveness of quantum-inspired fidelity as a foundation to improve reasoning trace coherence in LLMs.
\end{abstract}

\section{Introduction}
In recent times, large language models (LLMs) have made substantial progress in their ability to perform diverse natural language tasks such as open-domain question answering \cite{kamalloo2023evaluating} \cite{wang2024retrieve}, code generation \cite{jiang2024survey} \cite{wang2023review}, summarization \cite{subbiah2024reading}, and dialogue systems \cite{yi2402survey}. Despite the progress, their ability to perform coherent multi-step reasoning is limited even when the final answers are correct. The logical gaps in the underlying chains of thought pose challenges for applications in the fields of scientific discovery, legal analysis, and multi-hop factual verification where transparency, factual grounding, and coherent reasoning are essential. These shortcomings are often masked by fluent surface-level language but the fundamental pitfall with the current LLM alignment techniques remains. LLM alignment techniques such as supervised fine-tuning (SFT) \cite{li2023label}, Reinforcement Learning from Human Feedback (RLHF) \cite{gonzalez2025reinforcement}, Direct Preference Optimization (DPO) \cite{rafailov2023direct}, and Constitutional AI \cite{kyrychenko2025c3ai} optimize scalar reward signals or preference-based objectives that reflect end-task performance. While these approaches are effective at improving fluency, they don't always emphasize logical consistency and step-wise coherence within multi-step reasoning traces. 

To address the challenges with the current alignment approaches, this study proposes a novel quantum-inspired trace-alignment approach that precisely focuses on improving the internal consistency of multi-step reasoning traces. At the core of the proposed approach is a quantum-inspired Projected Entangled Pair States (PEPS) \cite{cirac2021matrix} model which draws from quantum many-body physics to capture entanglement across distributed systems. The PEPS formalism is used to represent reasoning traces as structured tensor networks, where each step in the trace is encoded as a projection entangled with the rest of the trace. By contracting the tensor network, the model computes the fidelity score, a scalar value that reflects the global consistency of the reasoning trace. The fidelity score serves as a holistic, structure-aware measure to determine the alignment of the individual steps as they form a coherent multi-step argument. However, the fidelity score is a non-differentiable sequence-level scalar signal that cannot be optimized directly using token-level objectives typically employed in approaches such as SFT. Therefore, this study employs a reinforcement learning (RL) formulation to incorporate the fidelity score into the model's behavior. Precisely, Proximal Policy Optimization (PPO) \cite{schulman2017proximal} is used as the policy optimization algorithm due to its robustness to noisy and high-variance rewards, and effectiveness in LLM fine-tuning. PPO allows the LLM to iteratively adapt its generation policy in response to the PEPS-guided feedback (fidelity score) while preserving training stability and avoiding policy collapse.

This work contributes to the research on quantum-inspired LLM alignment by applying PEPS, a tensor network formalism, to guide multi-step reasoning through RL. The contributions of the work are: 
\begin{itemize}
    \item The introduction of a quantum-inspired use case of PEPS to model multi-step reasoning as a structured tensor network to enable holistic coherence scoring via trace-level fidelity.
    \item The development of a PPO-based RL framework that optimizes multi-step reasoning using PEPS-based fidelity as a sequence-level reward signal, enabling direct optimization of logical consistency beyond token-level objectives. 
    \item The introduction of a comprehensive reasoning-aware evaluation framework that combines structural metrics with semantic similarity metrics to assess internal coherence and validity of multi-step reasoning traces.
\end{itemize}


\section{Related Work}

\textbf{Quantum-Inspired Tensor Networks for LLMs:} Recent advances in quantum-inspired modeling highlight the utility of tensor networks to enhance structured reasoning and drive architectural novelty. Kong et al. \cite{kong2025quantum} propose a Quantum Weighted Tensor Hybrid Network (QWTHN) that combines quantum neural circuits with tensor representations to achieve superior parameter-efficient fine-tuning compared to traditional Low-Rank Adaptation (LoRA) \cite{hu2022lora}, demonstrating improved generalization. Likewise, the work by Aizpurua, Jahromi, Singh, and Orus \cite{aizpurua2024quantum} integrates tensor network disentanglers and matrix product operators into transformer blocks, effectively decomposing dense weight matrices into structured, entangled components that capture richer contextual dependencies. Complementing these architectural innovations, Chen et al. \cite{chen2024quanta} propose QuanTA, a quantum-inspired tensor adaptation method that supports high-rank parameter updates using tensor operations inspired by quantum circuits. It enables expressive fine-tuning without additional inference overhead and outperforms LoRA variants on reasoning tasks. Ali, Delgado, and de Leceta \cite{ali2024quantum} survey the practical adoption of quantum-inspired tensor network 
 methods such as matrix product states MPS and PEPS in industrial and optimization contexts, highlighting their scalability and interpretability for AI applications. Krupp et al. \cite{krupp2025llm} evaluate the quality of LLM-generated tips in quantum computing education, finding that such tips can match or outperform expert-created ones in terms of conceptual helpfulness. Their findings reinforce the need for coherence-aware evaluation metrics such as tensor network–based fidelity scoring. These bodies of research collectively establish a solid foundation for leveraging tensor network principles such as entanglement and fidelity as a means of improving both structure and reasoning in language models.

\textbf{RL with Non-Traditional Rewards:} RL with LLMs has extended beyond the conventional supervised loss functions to incorporate non-traditional reward signals that reflect human preferences, factuality, and coherence. The foundational studies by Ziegler et al. \cite{ziegler2019fine} and Stiennon et al. \cite{stiennon2020learning} in RLHF introduce reward models trained on human comparisons to fine-tune LLMs using PPO, resulting in outputs more aligned with human preferences. Although RLHF is effective, it is limited by the scalability and sujectivity of human feedback. The study by Scheurer et al. \cite{scheurer2023training} proposes replacing human feedback with an LLM-based feedback generated by the LLM trained using imitation learning. Bai et al. \cite{bai2022constitutional} propose Constitutional AI, a framework that fine-tunes LLMs using artificial intelligence (AI) generated feedback based on a predefined set of ethical principles that reflect human preferences. Both \cite{scheurer2023training} and \cite{bai2022constitutional} demonstrate that LLMs can generate feedback on their own outputs, enabling scalable fine-tuning without relying on direct human supervision. Shmied et al. \cite{schmied2025llms} highlight the "knowing-doing" gap in LLMs where models exhibit correct reasoning trace, yet fail to act reliably on that knowledge. They propose RL fine-tuning on chain-of-thought outputs to align reasoning with action performance. Building on these advances, this study introduces a PEPS-based fidelity critic that evaluates the internal consistency of multi-step reasoning traces and provides a structured, scalable reward signal to guide RL (PPO) without relying on human annotations.

\textbf{Multi-Step Reasoning Evaluation:}  
The evaluation of multi-step reasoning traces in LLMs extends beyond measuring final-answer accuracy. Comprehensive assessment requires capturing properties such as logical consistency, contextual relevance, and semantic fluency, which recent studies address through different techniques. Entailment-based approaches by Canburu et al. \cite{camburu2018snli} assess groundedness and local validity by evaluating whether intermediate steps are logically supported by the input context. Logic-driven benchmarks introduced by Alfageeh et al. \cite{alfageeh2025prompts} evaluate reasoning behavior by mapping natural language prompts to formal logic representations. Ling et al. \cite{ling2023deductive} propose a decompositional verification framework that decomposes chain-of-thought reasoning into stepwise natural language deductions, enabling logical validation of intermediate inference.
While these approaches offer valuable insights, they are differ in scope, typically focusing on isolated properties such as factuality, validity, or structural consistency. The study by Lee and Hockenmaier \cite{lee2025evaluating} brings conceptual clarity to the evaluation space by proposing a taxonomy that categorizes reasoning quality into four dimensions: groundedness, validity, coherence, and utility. This study adopts the categorical taxonomy in its evaluation protocol to ensure that local step quality and global trace coherence are captured in a principled and annotation-free manner.

\section{Methodology}

\begin{figure*}[t]
    \centering
    \includegraphics[width=\textwidth]{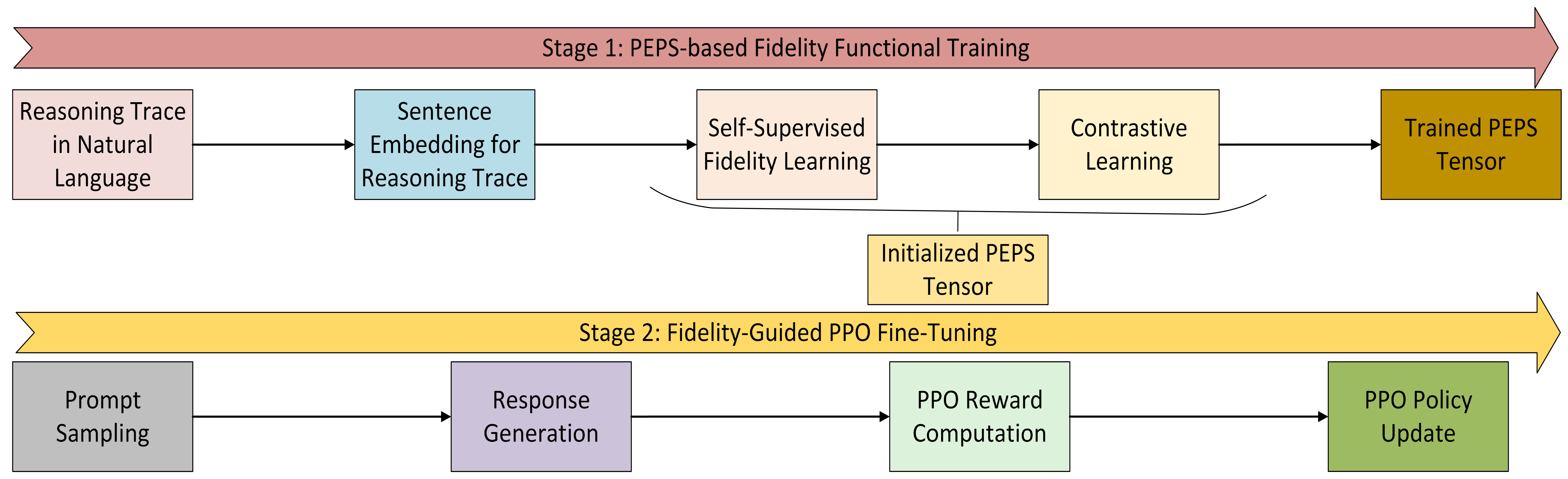}
    \caption{Overview of the Quantum-Inspired Two-Stage Training Pipeline}
    \label{fig:peps-method}
\end{figure*}

This study proposes a two-stage training pipeline to guide LLMs toward coherent multi-step reasoning. In the first stage, a quantum-inspired fidelity functional based on a PEPS tensor is trained to evaluate the structural coherence of reasoning traces. In the second stage, the pretrained fidelity functional is used as a reward signal to fine-tune a causal LLM using PPO-based RL. The two stages are illustrated in Figure \ref{fig:peps-method}, and the following subsections formalize each of the stages.

\subsection*{Stage 1: PEPS-based Fidelity Functional Training}
\subsection{Entangled Tensor Representations of Reasoning Traces}
\label{sec:Entagled_Tensor}
Let $\mathcal{X}$ and $\mathcal{R}$ denote the space of natural language queries and multi-step reasoning traces, respectively. For a given input query $x \in \mathcal{X}$, the LLM generates a reasoning trace $\mathbf{r} = (s_1, s_2, s_3, \dots, s_t) \in \mathcal{R}$, where each $s_t$ corresponds to a human interpretable intermediate reasoning step. The latent structure across these steps is formalized by embedding each \( s_t \) into a high-dimensional real Hilbert space \( \mathcal{H} \cong \mathbb{R}^d \), where \( \mathcal{H} \) is the semantic embedding space and \( \mathbb{R}^d \) is the \( d \)-dimensional space of real-valued sentence embeddings. This mapping is performed using a sentence-level encoder \( \Phi : \mathcal{V} \rightarrow \mathcal{H} \), where \( \mathcal{V} \) denotes the set of all valid token sequences over a fixed vocabulary. Each embedded step is denoted as $r_t = \Phi(s_t) \in \mathcal{H}$. The full reasoning trace is lifted into a tensor product space that encodes the semantic representation of the full reasoning trace across multiple reasoning steps. Each reasoning step $r_t$ is embedded into a local representation in $\mathcal{H}$, and the full trace, $\psi$, is mapped into the joint space $\mathcal{H}^{\otimes T}$ as:
\begin{equation}
    \psi = \Phi(\mathbf{r}) = r_1 \otimes r_2 \otimes \cdots \otimes r_T \in \mathcal{H}^{\otimes T}
\end{equation}

The tensor product captures the joint space of reasoning steps but not their correlations. A PEPS-based contraction integrates local semantics into a globally entangled representation, with coherence assessed through a fidelity functional, $\mathcal{F}_{\mathcal{T}}$. 

\subsection{PEPS Tensor Networks for Structural Coherence}
The fidelity functional, $\mathcal{F}_{\mathcal{T}}$, is parameterized by a PEPS tensor, $\mathcal{T} \in \mathbb{R}^{D \times D \times d \times D \times D}$ where the physical dimension $d$ corresponds to the semantic embedding space of each reasoning step $r_t \in \mathbb{R}^d$, and the four bond dimensions $D$ model latent pairwise interactions between adjacent reasoning steps. Shared across all positions in the reasoning trace, the PEPS tensor $\mathcal{T}$, serves as a latent contraction operator that integrates local semantics into a globally entangled structure, enabling coherence assessment through structured tensor contraction.

Each semantic embedding $r_t \in \mathbb{R}^d$ is incorporated into the PEPS-based tensor network through a mode-3 contraction (along semantic dimension) with the shared tensor $\mathcal{T}$, resulting in a structured chain of entangled representations across the reasoning steps:
\begin{equation}
\mathcal{C}^{[t]}(\mathcal{T}, r_t) = \sum_{k=1}^d \mathcal{T}_{\alpha\beta k \gamma\delta} (r_t)_k
\end{equation}
where $\mathcal{C}^{[t]}(\mathcal{T}, r_t) \in \mathbb{R}^{D \times D \times D \times D}$ is the contracted rank-4 PEPS tensor obtained upon the mode-3 contraction. The indices $\alpha, \beta, \gamma, \delta \in \{1, \dots, D\}$ represent the four latent variables that mediate pairwise interactions between adjacent reasoning steps, and the index $k \in \{1, \dots, d\}$ corresponds to the physical (semantic) dimension.

For a given entangled reasoning trace, $\psi$, its fidelity, $\mathcal{F}_{\mathcal{T}}(\psi)$, is computed by accumulating the \(\ell_2\)-norms of the contracted tensors:
\begin{equation}
\label{eq:fid_reward}
\mathcal{F}_{\mathcal{T}}(\psi) = \prod_{t=1}^{T} \left\| \mathcal{C}^{[t]}(\mathcal{T}, r_t) \right\|_2 
\end{equation}

The multiplicative form of the fidelity functional, $\mathcal{F}_{\mathcal{T}}$, ensure that local misalignments propagate globally, enforcing strict coherence across reasoning steps. If all local contractions yield high norms, the fidelity remains high, reflecting a globally consistent trace. However, by penalizing inconsistent steps that yield a low norm, the corresponding term proportionally reduces the overall score. This multiplicative aggregation mechanism appropriately amplifies the effect of inconsistencies, suitably rewards consistent alignment, and guides the model toward globally structured reasoning.

\subsection{Self-Supervised and Contrastive PEPS Tensor Training}
\label{sec:SSL_and_Contrastive_Tensor_Training}
Prior to RL-based fine-tuning using PPO, the PEPS tensor $\mathcal{T}$ is learned through a two-stage optimization process using a large corpus of structurally coherent, unlabeled multi-step reasoning traces, $\mathcal{D}_\text{traces}$. The first stage consists of self-supervised learning (SSL) wherein a given reasoning trace $\mathbf{r}^{(i)} \in \mathcal{D}_\text{traces}$ is embedded as $\psi^{(i)} = \Phi(\mathbf{r}^{(i)})$ and its coherence is evaluated using the fidelity functional, $\mathcal{F}_{\mathcal{T}}(\psi^{(i)})$. The training objective maximizes the log-fidelity, implemented as minimizing the negative log-fidelity loss, $\mathcal{L}_{\text{SSL}}$:
\begin{equation}
\mathcal{L}_{\text{SSL}}(\mathcal{T}) = - \frac{1}{N} \sum_{i=1}^{N} \log \mathcal{F}_{\mathcal{T}}(\psi^{(i)} + \varepsilon), \quad \varepsilon \footnote{The \(\varepsilon\) term is a small scalar constant that prevents instability from zero fidelity.}> 0
\end{equation}
Log-fidelity is chosen for optimization rather than raw fidelity to improve numerical stability during training. The use of log-fidelity converts the multiplicative aggregation of step-wise contractions into an additive form, mitigating the risk of numerical underflow that might arise from multiplying multiple values less than one. As a result, it provides smoother gradients especially for longer traces. Although this formulation is used for numerical stability, it naturally admits a quantum-inspired interpretation that parallels energy-based formulations in quantum many-body systems. The negative log-fidelity can be viewed as an energy potential over the latent Hilbert space of reasoning traces, where structurally coherent traces correspond to low-energy configurations and incoherent traces are assigned higher energy values.

After SSL, the PEPS tensor $\mathcal{T}$ is further refined using contrastive fine-tuning to enhance its ability to distinguish structurally coherent reasoning traces from corrupted or incoherent ones. This stage improves the discriminative sensitivity of the fidelity functional $\mathcal{F}_{\mathcal{T}}$ by explicitly optimizing it to assign higher scores to valid reasoning traces and lower scores to perturbed variants. 

Each original (coherent) reasoning trace from $\mathcal{D}_\text{traces}$ is paired with a non-coherent counterpart generated by augmenting the original reasoning trace using random shuffling of steps along with the substitution of semantically unrelated statements. The contrastive objective minimizes the relative fidelity of the non-coherent trace compared to its coherent counterpart using the loss function $\mathcal{L}_{\text{contrast}}(\mathcal{T})$, defined as:
\begin{equation}
    \mathcal{L}_{\text{contrast}}(\mathcal{T}) = - \log \frac{\mathcal{F}_{\mathcal{T}}(\psi^{+})}{\mathcal{F}_{\mathcal{T}}(\psi^{+}) + \mathcal{F}_{\mathcal{T}}(\psi^{-})}
\end{equation}

where $\psi^{+}$ and $\psi^{-}$ are the entangled representations of original (coherent) trace and corrupted trace respectively.

By explicitly contrasting between coherent and non-coherent reasoning traces, the contrastive learning complements the SSL objective to improve the overall fidelity critic before fine-tuning with PPO in Stage 2.

\subsection*{Stage 2: Fidelity-Guided PPO Fine-Tuning}
After SSL, the PEPS tensor \(\mathcal{T}\) is fixed and used as a fidelity-based reward critic to fine-tune a causal LLM \(\pi_\theta\) using PPO. Given an input query \(x\), the LLM samples a reasoning trace \(\psi \sim \pi_\theta(\cdot \mid x)\) and receives a composite reward, $R(\psi)$, that integrates structural coherence and novelty rewards. The composite reward is expressed as:
\begin{equation}
R(\psi) = \lambda_f \cdot \log \mathcal{F}_{\mathcal{T}}(\psi) 
+ \lambda_r \cdot \mathcal{H}_{\text{novelty}}(\psi), \quad \lambda_f + \lambda_r = 1
\end{equation}   
where:
\begin{itemize}
    \item \( \mathcal{F}_{\mathcal{T}}(\psi) \) (eq. \ref{eq:fid_reward}) is the PEPS fidelity score measuring global structural coherence of the reasoning trace. 

    \item \( \mathcal{H}_{\text{novelty}}(\psi) \) is the \(n\)-gram-based novelty reward, given by the proportion of unique \(n\)-grams to the total number of \(n\)-grams, given by: 
    \begin{equation}
\mathcal{H}_{\text{novelty}}(\psi) = \frac{ \left| \left\{ g_i \right\}_{i=1}^{K} \right| }{K}
\end{equation}
where \(g_i\) is the \(i\)-th \(n\)-gram,  and \(K = L - n+1 \) where \(L\) is the total number of tokens across the steps of the reasoning trace \(\psi\).

\end{itemize}

This composite reward is intentionally curated to reinforce complementary aspects of high-quality reasoning. The first term, \( \log \mathcal{F}_{\mathcal{T}}(\psi) \), serves as the structural core of the reward, quantifying the global coherence of the reasoning trace to ensure that the sequence of reasoning steps adheres to entanglement-informed patterns of logical consistency learned in Stage 1. The second term, \( \mathcal{H}_{\text{novelty}}(\psi) \), encourages informational diversity by rewarding reasoning traces that exhibit low semantic redundancy, thereby promoting progression and variation across the reasoning trace. Collectively, these two terms provide a well-calibrated reward that supports structural coherence, semantic relevance, and step-level diversity in a unified and interpretable manner.

The LLM is fine-tuned using PPO which updates the LLM by maximizing a clipped surrogate objective:
\begin{equation}
\begin{aligned}
\mathcal{L}_{\text{PPO}}(\theta) = \mathbb{E}_{\psi} \Big[ 
& \min\big( \rho(\psi) A(\psi), \\
& \quad \text{clip}(\rho(\psi), 1 - \epsilon, 1 + \epsilon) A(\psi) \big) 
\Big] + \beta \mathcal{H}[\pi_\theta]
\end{aligned}
\end{equation}
where \( \theta \) denotes the parameters of the LLM policy \( \pi_\theta \), \( \mathbb{E}_\psi \) represents the expectation over sampled reasoning traces \( \psi \sim \pi_\theta(\cdot \mid x) \), \( \rho(\psi) = \frac{\pi_\theta(\psi)}{\pi_{\theta_{\text{old}}}(\psi)} \) is the likelihood ratio between the current and previous policy, and \( A(\psi) = R(\psi) - \hat{V}(x) \) is the advantage function based on the difference between the obtained reward, \(R(\psi)\), and a learned value baseline, $\hat{V}(x)$, approximated by an exponential running mean. The hyperparameter \( \epsilon \) controls the clipping range for the policy update, and \( \beta \) weights the contribution of the entropy term. The policy entropy is defined as:
\begin{equation}
\mathcal{H}[\pi_\theta] = -\sum_{t} \sum_{w \in \mathcal{V}} \pi_\theta(w_t \mid h_t) \log \pi_\theta(w_t \mid h_t)
\end{equation}
where \( \mathcal{V} \) is the vocabulary and \( h_t \) denotes the decoder state at time step \( t \). 

Two auxiliary terms are incorporated into the PPO objective to regularize optimization and prevent the model from collapsing:
\begin{itemize}
    \item Kullback-Leibler (KL) penalty, $\mathcal{L}_{\text{KL}}$, to ensure that the updated policy $\pi_\theta$ does not deviate excessively from a reference policy $\pi_{\text{ref}}$, defined as:
    \begin{equation}
        \mathcal{L}_{\text{KL}} = D_{\text{KL}}\big( \pi_\theta(\cdot \mid x) \, \| \, \pi_{\text{ref}}(\cdot \mid x) \big)    
    \end{equation}
    \item Supervised auxiliary loss, $\mathcal{L}_{\text{sup}}$, to encourage the model to stay consistent with gold reasoning traces $y^*$ by minimizing the negative log likelihood of the correct tokens, defined as:
    \begin{equation}
\mathcal{L}_{\text{sup}} = \mathbb{E}_{x, y^*} \left[ -\sum_{t} \log \pi_\theta(y^*_t \mid y^*_{<t}, x) \right]    
    \end{equation}
    
\end{itemize}

The final PPO optimization objective, $\mathcal{L}_{total}$, is given as:
\begin{equation}
    \mathcal{L}_{total} = \mathcal{L}_{\text{PPO}} + \beta_{\text{KL}} \cdot \mathcal{L}_{\text{KL}} + \lambda_{\text{sup}} \cdot \mathcal{L}_{\text{sup}}    
\end{equation}

where $\beta_{\text{KL}}$ and $\lambda_{\text{sup}}$ are adjustable weighting coefficients to control the auxiliary terms. 

PPO is chosen as the fine-tuning algorithm due to its stability and efficiency in optimizing complex, non-differentiable reward functions such as the PEPS-based fidelity signal. Unlike supervised objectives, the PEPS reward does not decompose well over tokens, making gradient propagation challenging. PPO's clipped objective allows controlled updates to the LLM policy in response to these structured, global rewards. By integrating the PEPS-based fidelity signal with semantic similarity and repetition penalties, it ensures that coherence-promoting feedback is absorbed without destabilizing the LLM's behavior. This makes PPO well-suited in the study's quantum-inspired setting where small deviations in trace structure might lead to large changes in fidelity. Therefore, the combination of PPO and PEPS supports consistent, step-aligned, and globally entangled reasoning traces. 

\section{Experiment}
\subsection{Research Questions}
\label{sec:research-questions}

This study is motivated by the following research questions (RQ):

RQ1: Does PPO with PEPS-based fidelity rewards improve structural coherence in multi-step reasoning traces compared to baseline approaches without structural rewards?

RQ2: Do PEPS-guided PPO policies produce reasoning traces that are more semantically aligned with the query compared to baseline approaches that generate reasoning traces without structural rewards?

RQ3: Do structurally guided PPO policies generalize better across diverse reasoning tasks compared to baseline approaches without structural rewards?

\subsection{Datasets}
\label{sec:datasets}
The proposed quantum-inspired approach is evaluated on three benchmark datasets, EntailmentBank, StrategyQA, and Grade School Math 8K (GSM8K). All three datasets offer question-answer pairs accompanied by human-authored, multi-step gold reasoning traces. Each of the datasets is described as follows.

The EntailmentBank dataset \cite{entailmentbank2021} consists of 1,840 science and general knowledge questions sourced from the WorldTree V2 corpus, each paired with a hypothesis and its corresponding entailment chain. Each question is annotated with an annotation tree, a structured multi-step reasoning trace where intermediate conclusions are derived from a combination of supporting facts and prior inferences. The leaf nodes represent atomic facts and internal nodes represent intermediate conclusions. Gold-standard reasoning traces are offered for the dataset that make it suitable for evaluating coherence and fidelity in multi-step reasoning.

The StrategyQA dataset \cite{Geva2021DidAU} consists of 2,780 yes/no questions with gold-standard answer annotations and human-authored supporting facts. Each question is designed to require implicit reasoning across multiple knowledge steps, which are not directly stated in the question. The human-authored multi-step justifications articulate these latent connections and serve as high-quality traces for model training and evaluation.

The GSM8K dataset \cite{cobbe2021gsm8k} comprises 8,500 questions pertinent to elementary school mathematics wherein each question requires multiple steps of numerical reasoning to arrive at the final answer. Each of the questions is written in natural language and accompanied by gold-standard solutions that break down the reasoning process into multiple steps. Its natural language questions and multi-step gold solutions enable fine-grained evaluation of structural coherence for mathematical reasoning tasks. 


\subsection{Model Training}
The pretrained TinyLLaMA-1.1B model \cite{zhang2024tinyllama} is used as the base causal LLM to facilitate computationally efficient training and broader hyperparameter exploration. This model retains architectural alignment with the larger instruction-tuned LLaMa models such as LLaMA2-7B and LLaMA2-13B, making it representative of decoder-style transformers. Its compact size allows for the experiment to be conducted on academic research-class Nvidia RTX Ada 6000 GPU, representative of high-performance setups accessible outside large-scale datacenter infrastructure. Finally, the LLaMA family of LLMs is open-access which aids the reproducibility of the experiment even in financial resource-constrained environments.

The model for the proposed quantum-inspired approach is trained wherein, each dataset is split into 40\% for Stage 1 PEPS pretraining, 45\% for PPO finetuning, and 15\% held out for exclusively for evaluation. This partitioning prevents overlap between the data used for structural learning in Stage 1 and policy fine-tuning in Stage 2, ensuring that evaluation reflects true generalization. The hyperparameters used for Stage 1 and Stage 2 are as shown in Table \ref{tab:stage1-hparams} and Table \ref{tab:stage2-hparams} respectively.

\begin{table}[h]
\centering
\caption{Stage 1: PEPS Pretraining Hyperparameters}
\label{tab:stage1}
\begin{tabular}{l l}
\toprule
\textbf{Hyperparameter} & \textbf{Value} \\
\midrule
Embedding model & all-MiniLM-L6-v2 \\
Embedding dimension ($d$) & 384 \\
Bond dimension ($D$) & 30 \\
Optimizer & Adam \\
Learning rate & $1 \times 10^{-3}$ \\
Epochs (SSL/Contrastive) & Dynamic (early stopping) \\
Early stopping patience & 2 epochs \\
\bottomrule
\end{tabular}
\label{tab:stage1-hparams}
\end{table}

\begin{table}[h]
\centering
\caption{Stage 2: PPO Fine-Tuning Hyperparameters}
\begin{tabular}{lc}
\toprule
\textbf{Hyperparameter} & \textbf{Value} \\
\midrule
Optimizer & AdamW \\
Learning Rate & $5 \times 10^{-6}$ \\
Steps per Epoch & 100 \\
Fidelity Coefficient ($\lambda_f$) & 0.8 \\
Novelty Coefficient ($\lambda_r$) & 0.2 \\
KL Penalty Coefficient ($\beta_{\text{KL}}$) & 0.1 \\
Supervised Loss Coefficient ($\lambda_{\text{sup}}$) & 0.3 \\
n-gram size (n) & 3 \\
Entropy Coefficient ($\beta$) & 0.01 \\
\bottomrule
\end{tabular}
\label{tab:stage2-hparams}
\end{table}

\subsection{Evaluation Metrics}
\label{sec:evaluationmetrics}
Table \ref{tab:metrics-summary} summarizes the evaluation metrics along with their dimension, computational formulations and underlying models.

\begin{table*}[h]
\small
\centering
\caption{Evaluation metrics categorized by reasoning quality}
\label{tab:metrics-summary}
\begin{tabular}{lll p{4.2cm} p{3.5cm}}
\toprule
\textbf{Dimension} & \textbf{Metric} & \textbf{Model Used} & \textbf{Description} & \textbf{Formula} \\
\midrule

\multirow{2}{*}{Groundedness} 

& Mean Entailment Confidence (MEC)
& DeBERTa-MNLI 
& Computes the average entailment probability between the input question and generated response. Reflects how strongly the response is supported (entailed) by the input without thresholding.
& \( \frac{1}{N} \sum_{i=1}^{N} P_i^{\text{entail}} \) where $P_i^{\text{entail}}$ is the entailment probability of the $i^\text{th}$ question-response pair. \\

\midrule

Validity 
& Weighted Entailment Score (WES)
& DeBERTa-MNLI 
& Assesses the logical consistency of a reasoning trace by measuring how well each step follows from the prior context. This method gives partial credit to steps that are logically neutral, capturing strong logical connections and non-contradictory transitions. 
& \( P_i^{\text{weighted}} = 1.0 \cdot P_i^{\text{entail}} + 0.5 \cdot P_i^{\text{neutral}} \) where \( P_i^{\text{entail}} \) and \( P_i^{\text{neutral}} \) are entailment and neutrality probabilities respectively, for step \( i \) based on prior context and current reasoning step. \\

\midrule

\multirow{2}{*}{Coherence} 
& BERT Score (BERT)
& Sentence-BERT 
& Measures embedding-based cosine similarity between generated and reference outputs to capture surface-level reasoning trace coherence.
 
& \( \cos(\mathbf{e}_{\text{gen}}, \mathbf{e}_{\text{ref}}) \) \newline where \( \mathbf{e}_{\text{gen}} \) and \( \mathbf{e}_{\text{ref}} \) are embedding vectors of generated and reference outputs. \\

\midrule

Utility 
& BLEURT Score (BLEURT)
& BLEURT-20 
& Measures semantic similarity between the generated and reference outputs using a learned regression model fine-tuned on human-rated examples. 
& \( \text{BLEURT}(x, y) = f_\theta(\text{BERT}(x, y)) \) where \( x \) is the reference, \( y \) is the generated response, and \( f_\theta \) is the regression head. \\

\bottomrule
\end{tabular}
\end{table*}

\subsection{Baselines and Results}
\label{sec:Baselines_and_Results}

\begin{figure*}[t]
    \centering
    \includegraphics[scale=0.40]{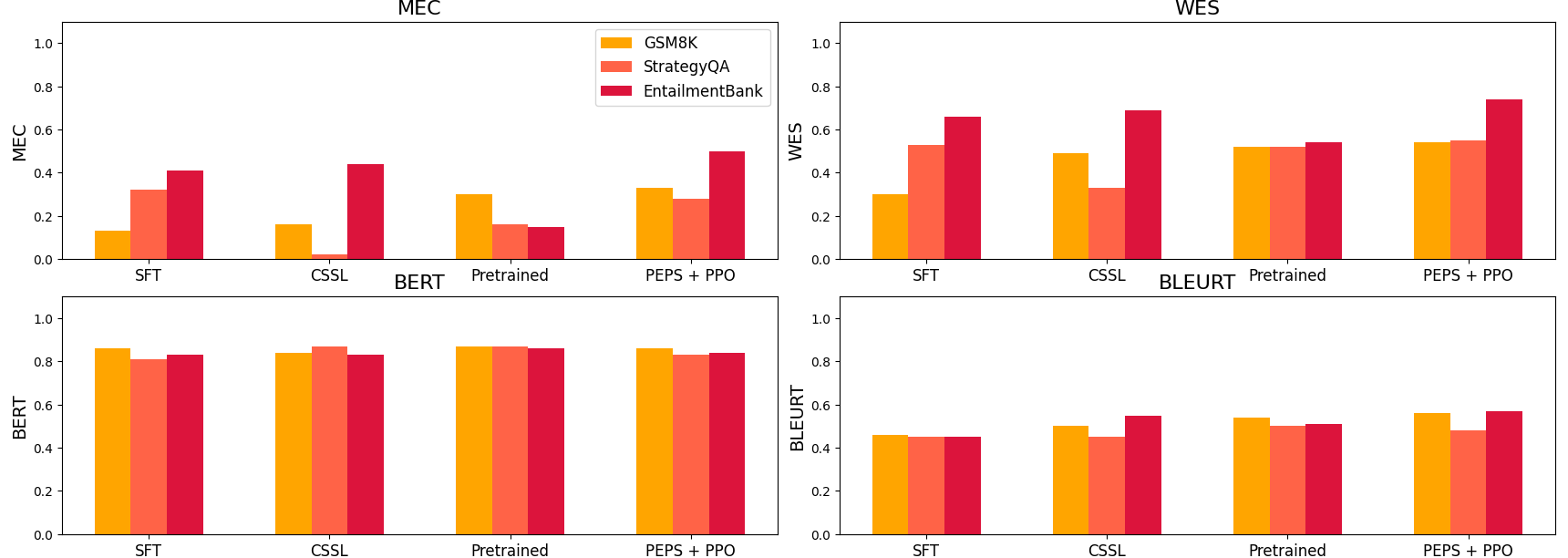}
    \caption{Evaluated Metrics by Technique on the GSM8K, StrategyQA, and EntailmentBank Datasets}
    \label{fig:metrics}
\end{figure*}

The effectiveness of the proposed quantum-inspired approach is benchmarked by comparing with three different baseline approaches: 
\begin{itemize}
    \item Inference on the pretrained TinyLLaMa-1.1B model (Pretrained) without any fine-tuning.
    \item A supervised fine-tuning (SFT) baseline wherein the pretrained TinyLLama-1.1B model is trained on 85\% of the dataset with the rest held out for testing.
    \item A contrastive SSL (CSSL) baseline with TinyLLaMa-1.1B which uses cosine margin loss to distinguish coherent reasoning traces from non-coherent traces. The non-coherent traces are generated using the same corrupt trace generation strategy outlined for contrastive fine-tuning in section \ref{sec:SSL_and_Contrastive_Tensor_Training}. A train-test split of 85-15\% is used for this approach.
\end{itemize}
These baselines span from zero-shot evaluation to task-specific learning, offering a well-rounded comparison with the structured reward-driven fine-tuning using the proposed approach. The mean of the results\footnote{Results are reported on the same test set to ensure evaluation consistency.} for the proposed approach (PEPS+PPO), and the baselines on each of the datasets are shown in Table \ref{tab:technique_comparison}. Figure \ref{fig:metrics} offers a visual comparison of the evaluation metrics by dataset.

\begin{table*}[htbp]
\centering
\caption{Comparison of Techniques Across Datasets using Four Evaluation Metrics}
\label{tab:technique_comparison}
\renewcommand{\arraystretch}{1.3}
\setlength{\tabcolsep}{8pt}
\begin{tabular}{|l|cccc|cccc|cccc|}
\hline
\multirow{2}{*}{\textbf{Technique}} 
& \multicolumn{4}{c|}{\textbf{GSM8K}} 
& \multicolumn{4}{c|}{\textbf{StrategyQA}} 
& \multicolumn{4}{c|}{\textbf{EntailmentBank}} \\
\cline{2-13}
& \textbf{MEC} & \textbf{WES} & \textbf{BERT} & \textbf{BLEURT} 
& \textbf{MEC} & \textbf{WES} & \textbf{BERT} & \textbf{BLEURT} 
& \textbf{MEC} & \textbf{WES} & \textbf{BERT} & \textbf{BLEURT} \\
\hline
SFT        &   0.13    &  0.3     &  0.86     &  0.46     &  \textbf{0.32}     &  0.53     &  0.81     &  0.45     &    0.41   &  0.66     &  0.83     &  0.45     \\
CSSL       &   0.16    &    0.49   &   0.84    &   0.5    &   0.02    &   0.33    &   \textbf{0.87}    &   0.45    & 0.44      &  0.69     &    0.83   &   0.55    \\
Pretrained       &   0.30    &  0.52    &  \textbf{0.87}     &  0.54     &  0.16     &  0.52     &  \textbf{0.87}     & \textbf{0.50}      &  0.15     &  0.54     &   \textbf{0.86}    &  0.51     \\
PEPS + PPO         &   \textbf{0.33}    &  \textbf{0.54}     &   0.86    &  \textbf{0.56}     &  0.28     &  \textbf{0.55}     &   0.83    &  0.48     &   \textbf{0.50}    &  \textbf{0.74}     & 0.84      &  \textbf{0.57}     \\
\hline
\end{tabular}
\end{table*}

\section{Discussion}
\label{sec:Discussion}
RQ1: Does PPO with PEPS-based fidelity rewards improve structural coherence in multi-step reasoning traces compared to baseline approaches without structural rewards?

The MEC and WES metrics are used to assess structural coherence as they quantify how strongly a reasoning step is entailed by original query (MEC), and how consistently the trace maintains neutrality throughout (WES). On GSM8K, where multi-step decomposition of arithmetic problems is critical, PEPS+PPO achieves MEC and WES of 0.33 and 0.54 respectively, outperforming the best baseline (Pretrained: MEC 0.3, WES 0.52) by 10\% in MEC and 3.9\% in WES suggesting that PEPS-based fidelity rewards capture a more complete sequence of intermediate computations while maintaining coherent solution steps. On EntailmentBank, where reasoning is framed as entailment chains over multiple inferential steps, PEPS+PPO achieves the strongest performance (MEC 0.5, WES 0.74), improving over the strongest baseline (CSSL: MEC 0.44, WES 0.69) by 13.6\% in MEC and 7.3\% in WES, indicating stronger ability to capture hierarchical entailment structures. On StrategyQA, where most questions rely on relational knowledge rather than step-by-step decomposition, the improvements offered by PEPS+PPO are more modest. PEPS+PPO achieves a WES of 0.55, surpassing the best baseline (SFT: 0.53) by 3.8\%, while the MEC is at 0.28, a  12.5\% decrease in MES relative to the best baseline (SFT: 0.32). It is worth mentioning that PEPS+PPO still outperforms two other baselines in MEC (CSSL: 0.02 and Pretrained: 0.16). While gains vary depending on task characteristics, the results demonstrate that the proposed approach offers consistent advantages in tasks emphasizing structured multi-step reasoning, while remaining competitive across diverse reasoning formats. 

RQ2: Do PEPS-guided PPO policies produce reasoning traces that are more semantically aligned with the query compared to baseline approaches that generate reasoning traces without structural rewards?

The WES and BLEURT metrics are used to measure the semantic alignment of the generated reasoning trace with the original query. WES rewards logical entailment, while BLEURT captures sentence-level fluency and semantic similarity in alignment with human judgment. On GSM8K, PEPS+PPO achieves a WES of 0.54 and BLEURT of 0.56, improving over the best baseline (Pretrained: WES 0.52, BLEURT 0.54) by 3.9\% in WES and 3.7\% in BLEURT, indicating better alignment of intermediate reasoning steps and human-aligned semantic correspondence on a dataset focused on step-wise arithmetic. On EntailmentBank, where semantic alignment involves composing intermediate entailments that collectively support the target conclusion, PEPS+PPO, with a WES of 0.74 and BLEURT of 0.57, outperforms the best baseline (CSSL: WES 0.69, BLEURT 0.55) by 7.3\% in WES and 3.6\% in BLEURT, suggesting that PEPS-guided policies generate more semantically coherent entailment chains. On StrategyQA, where questions involve broader relational knowledge, PEPS+PPO achieves a WES of 0.55 bettering the best baseline (SFT: 0.53) by 3.7\%, while showing a marginal decrease of 4\% in BLEURT with 0.48, relative to Pretrained (BLEURT: 0.5). Overall, the results suggest that PEPS-guided PPO policies offer consistent improvements in semantic alignment for tasks that emphasize structured reasoning, while remaining competitive in tasks with less explicit structure.

RQ3: Do structurally guided PPO policies generalize better across diverse reasoning tasks compared to baseline approaches without structural rewards?

The generalization of the proposed PEPS+PPO approach is analyzed across tasks that differ in both reasoning structure and linguistic form, using complementary evaluation metrics. A consistent pattern emerges, with baseline models often achieving isolated peaks on individual metrics, whereas PEPS+PPO maintains a stable balance across structural (MEC, WES) and semantic (BERT, BLEURT) dimensions. On datasets requiring explicit multi-step reasoning, such as GSM8K and EntailmentBank, PEPS+PPO achieves strong gains in the structural metrics of MEC (+10\% on GSM8K and +13.6\% on EntailmentBank) and WES (+3.9\% on GSM8K and +7.3\% on EntailmentBank), while also improving semantic alignment measured by BLEURT (+3.7\% on GSM8K and +3.6\% on EntailmentBank). On Strategy QA, which focuses more on associative knowledge than stepwise decomposition, PEPS+PPO remains competitive, with WES improving decently (+3.7\%) and BLEURT and MEC declining (-4\% in BLEURT and -12.5\% in MEC) over the best baseline. Although performance on surface-level similarity metrics such as BERT exhibits minor reductions across datasets (GSM8K: -1.1\%, EntailmentBank: -2.3\%, StrategyQA: -4.6\%) over the best baselines, PEPS+PPO consistently achieves high absolute BERT across all tasks. These trends suggest that while PEPS-guided policies prioritize structural coherence during reasoning, they still preserve contextual semantic alignment.

Another key observation among the baselines is there isn't a singular baseline that outperforms the rest on all datasets. Baseline approaches that perform well on one dataset exhibit performance degradation on others, reflecting sensitivity to dataset-specific reasoning patterns. In contrast, PEPS+PPO exhibits greater stability across diverse reasoning tasks, likely due to the inductive bias introduced by structural fidelity rewards. This structural regularization encourages the model to construct globally coherent reasoning chains, which are feasible across different tasks that differ in reasoning depth, decomposition, and knowledge demands.

\FloatBarrier

\section{Future Work and Conclusion}
Although the results demonstrate that the PEPS-guided PPO policies improve structural coherence and semantic alignment across diverse reasoning tasks, a few limitations remain. The PEPS fidelity objective emphasizes structural grounding but remains agnostic to semantic correctness, leading to variable gains on semantically underspecified datasets such as StrategyQA. Furthermore, the approach does not model factual consistency, meaning that coherent reasoning traces may still contain semantically incorrect or hallucinated content. Future work will pursue hybrid rewards integrating structural fidelity with semantic and factual constraints, alongside human-in-the-loop refinement. Overall, the study highlights quantum-inspired RL as a promising path for enhancing reasoning coherence in LLMs by bridging symbolic structure and neural representations.

\section*{Acknowledgment}
This work is supported by the Pollinator Health: Research and Application no. 2024-67014-42301 from the U.S. Department of Agriculture’s National Institute of Food and Agriculture.

\bibliography{references.bib}

\end{document}